# Decision Tree J48 at SemEval-2020 Task 9: Sentiment Analysis for Code-Mixed Social Media Text (Hinglish)

**Gaurav Singh**
University of Leeds
School of Computing
`sc19gs@leeds.ac.uk`

## Abstract

This paper discusses the design of the system used for providing a solution for the problem given at SemEval-2020 Task 9 where sentiment analysis of code-mixed language Hindi and English needed to be performed. This system uses Weka as a tool for providing the classifier for the classification of tweets and python is used for loading the data from the files provided and cleaning it. Only part of the training data was provided to the system for classifying the tweets in the test data set on which evaluation of the system was done. The system performance was assessed using the official competition evaluation metric F1-score. Classifier was trained on two sets of training data which resulted in F1 scores of 0.4972 and 0.5316.

## 1 Introduction

The SemEval-2020 Task 9 was about performing sentiment analysis on Code-Mixed Social Media Text which is mixture of two languages Hindi and English. The data contained in this code-mixed text was presented as tweets written in Code-mixed language comprising of Hindi and English. Code-Mixed language is a language in which people write text using mix of two or more languages, where some words belong to one language vocabulary and some words belong to another language vocabulary (CodaLab, 2017). This code-mixing is usually practiced by people who are bilingual or multilingual that is people who know two or more languages. Sentiment analysis is process where the sentiments behind a text written by people are analyzed using an automated process. The sentiments behind a written text can be categorized as negative, neutral, positive, or angry, sad, happy or strongly negative, negative, neutral, positive, strongly positive depending on the choice of the analysis (MonkeyLearn, 2020). An example of code-mixed tweet written in Hinglish with positive sentiment polarity is "all india nrc lagu kare w kashmir se dhara ko khatam kare ham indian ko aps yahi umid hai" – positive (Patwa et al, 2020).

SemEval (Semantic Evaluation) is a series of competitions held yearly for evaluating computational semantic analysis systems performing semantic analysis on various languages. SemEval is evolved from Senseval competition series which is mostly focussed on finding out the sense of a word with respect to the context in which it is written. Senseval competitions involved assessing word sense disambiguation algorithms (Wikipedia contributors, 2020). SemEval-2020 competition is a part of SemEval competition series held in year 2020 available at `http://alt.qcri.org/semeval2020/index.php?id=tasks`. All the tasks of the SemEval competitions were held on CodaLab website available at `https://codalab.org/`.

## 2 Data Description

The data was provided by the organizers at CodaLab website available at `https://competitions.codalab.org/competitions/20654#participate` for download through separate links for trial data, training data, validation data, and the test data. After the competition ended one more link was provided by the organizers for the download of the test labels.

Data was provided in the tab separated format in the files which had .txt as extensions. In the files, each tweet was split up into the corresponding tokens which could be a word or a universal symbol where each token was present in a different line and lang_id for each word was provided along with it separated by a tab which were Eng, Hin and O where Eng denoted that the word belongs to English dictionary and Hin denoted that the word belongs to Hindi dictionary and O denoted that the word is a universal symbol. In the starting of every tweet meta data of that tweet is provided in a line containing meta keyword followed by a unique number which denoted its unique identifier number separated by tab followed by sentiment polarity of that tweet again separated by tab which could be negative, neutral or positive.

## 3 Tools Used

Tools used for the classification of the data were Weka, Python and Jupyter Notebook. Python was used for initial loading of the data from the files downloaded and for cleaning of the data to make it fit for training of the classifier. Python was practiced in Jupyter Notebook. Weka was used for the classification purpose, where a classifier was chosen and trained with the data available and the results were obtained on the test data.

### 3.1 Weka

Waikato Environment for Knowledge Analysis (Weka) is a free software developed by the University of Waikato, New Zealand. It is available for free and is licensed under the GNU General Public License (Wikipedia contributors, 2020). Weka is an interactive tool with graphic user interface which has functionalities for data visualization and various machine learning algorithms for classification purposes. Weka is built using java so it can run on any machine having java installed on it.

### 3.2 Python

Python is an open source, high level language which supports object oriented paradigm and can be used to write general program and software. Python can be downloaded from `https://www.python.org/downloads/` and can be installed on any machine and used for programming. Python can also be installed by installing the anaconda distribution of python which is also open source software which contains several integrated development environment for practicing the python language. Many libraries have been developed for python programming which can be used for various data visualization, data analytics and machine learning algorithms.

### 3.3 Jupyter Notebook

Jupyter Notebook is an open source application developed as an integrated development environment for running the python codes. Jupyter Notebook requires python to be installed first for it to run. Jupyter Notebook is an acronym of the names of the programming languages it supports which are Julia, python and R. Jupyter Notebook has the functionality of cells which can allow the execution of a small snippet of code instead of running the whole file. All the python code was run in Jupyter Notebook.

## 4 Data Preparation

Data preparation tasks were performed by creating the codes using the python programming language in the Jupyter Notebook. Pandas library of the python language was used mainly for all the tasks of the data preparation. Data was first consolidated and made in a format that made it appear more suitable than the raw format. Data was then cleaned and only necessary things were kept and unwanted text was removed which constituted noise. After preparation, data was put into the files suitable for uploading into the weka which are in the arff format.

### 4.1 Data Consolidation

Raw data was given in the form of words and symbols present in separate lines which was consolidated to combine the words to make sentences for the individual tweets. This was done by loading the data into the dataframes data structures provided by the pandas library of python. After converting, tweets were stored along with their uid and sentiment polarity in the dataframes. This process was done with the training, validation and test data and separate files containing the consolidated data were created for them. The following image shows the snippet of code used for this process.

```
In [60]: text=""
         cla=""
         cla=df['class'][0]
         j=df['type'][0]
         for i in range(1,len(df)):
             if(df['word'][i]=="meta"):
                 df1 = df1.append({"ID": j, "Text": text, "class": cla}, ignore_index=True)
                 cla=df['class'][i]
                 j=df['type'][i]
                 text=""
             else:
                 text += " " + str(df['word'][i])
         df1 = df1.append({"ID": j, "Text": text, "class": cla}, ignore_index=True)
```

Figure 1: Data Consolidation.

## 4.2 Data Cleaning

Data Cleaning was done by executing the code in the python language which loaded the data and cleaned it using predefined libraries of python in Jupyter Notebook. Consolidated Data was loaded into the dataframes data structures provided by the pandas library of the python and then data was cleaned by building a small snippet of code which removed numbers, hash tags and any kind of special character from the data. First @User mentions were removed from the code using the regular expressions library. Then html tags were removed and html decoding which are things like " > were converted to actual symbols using the beautifulsoup library of the python. Then characters which are not alphabets were removed from the text. These data cleaning tasks were performed by importing library for regular expression and beautiful soup. Screenshot of the code used for data cleaning is depicted below.

```
In [14]: import re
         from bs4 import BeautifulSoup
         def clean_tweet(tweet):
             text = re.sub(r'@[A-Za-z0-9]+','', tweet)  # remove mentions [like @User]
             text = BeautifulSoup(text, 'lxml').get_text() # remove html tags and converts html decoding [things like ", >
             text = re.sub("[^a-zA-Z]", " ", text) # only keeps in letter characters [removes numbers, hash tags and other specia

             return text
```

Figure 2: Data Cleaning.

## 4.3 Creating Arff Files

Arff (Attribute-Relation File Format) files suited for uploading in the weka for performing the data modelling tasks were also created by executing a code in the python programming language in the Jupyter Notebooks. First an empty file was created with the write operation using the data handling functions of python. Then the relation and attribute instances were defined in the file. Then data keyword was declared following which cleaned data from the dataframes was appended in the file using the append mode of writing data from the dataframes in the file. Following image shows the code used for creating the arff files.

```
In [96]: f=open('/Users/gaurav/Desktop/conll/training_data_cleaned1.arff','w')
         f.write("@relation twitter\n
         @attribute id numeric\n
         @attribute tweet string\n
         @attribute subtask_a {neutral, negative, positive}\n
         @data\n")
         f.close()
         df1.to_csv("/Users/gaurav/Desktop/conll/training_data_cleaned1.arff",mode='a',header=False,index=False)
```

Figure 3: Creating Arff Files.

## 5 Modelling

Weka was used for selecting the model for training. Filtered Classifier was selected as the classifier for the classification of tweets. Filtered classifier provides the option of choosing a filter through which data is first processed before it is passed for training the model. It also provides the option of choosing any arbitrary model for the classification purposes. String to Word Vector filter was chosen as the filter used for features extraction from the text data. J48 Decision Tree was selected as the model used for the classification of tweets in the test data after training the model with the training data. This filtered classifier is present under the meta category of the classifiers in weka. Other than choosing the filter and the classifier, all the options were kept as default in the Filtered Classifier. The following image shows the options chosen for the classifier.

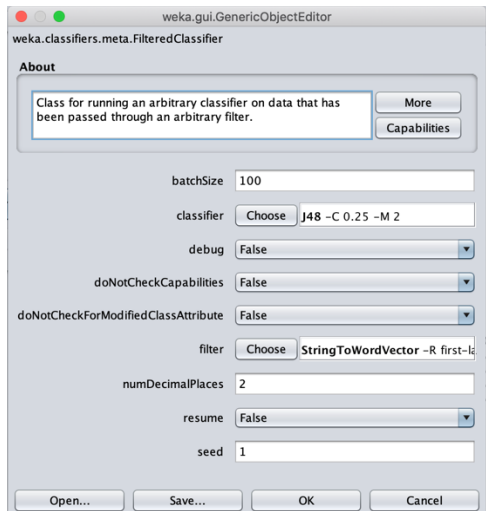
Figure 4: Filtered Classifier

## 5.1 Transforming Data (Features Extraction)

String To Word Vector Filter was chosen as the filter for extracting the features out of the string data. String to word vector filter converts each token in the string which are words separated by space into the vectors of numbers based on the parameters specified for converting the strings into numeric attributes. This vector of numbers represent the occurrence of words in the string. All the attributes of the string to word vector filter were kept as default. By default, string to word vector filter denotes the presence or absence of the word in a string using a binary number which is known as one hot encoding.

## 5.2 J48 Decision Tree

J48 Decision tree was chosen as the model for training the classifier. J48 is the name of the decision tree algorithm used in weka for classification. It uses release 8 of C4.5 algorithm for making the decision trees (Sefik Ilkin Serengil, 2018). C4.5 algorithm is an extended version of the Ross Quinlan's earlier version of algorithm for building the decision trees known as ID3 algorithm which is also developed by Ross Quinlan (Wikipedia contributors, 2020). So, basically J48 is an improved version of earlier version of ID3 algorithm which overcomes some of its shortcomings. The improved features implemented in C4.5 algorithm are working with missing data, working with discrete and continuous data and pruning to avoid overfitting (Sumit Saha, 2018). The default parameter values for J48 algorithm were kept for training the model with the training data. The following two images shows the screenshots of the attributes kept for the String To Word Vector filter and J48 decision tree respectively.

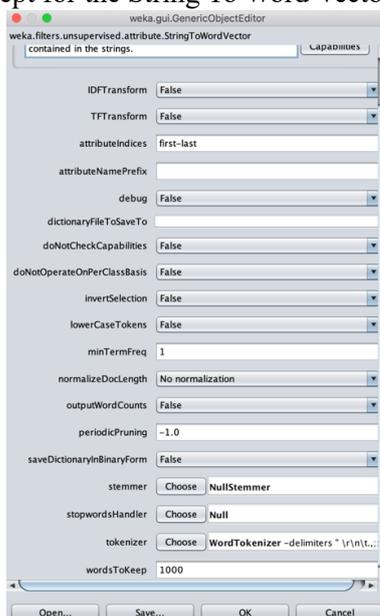
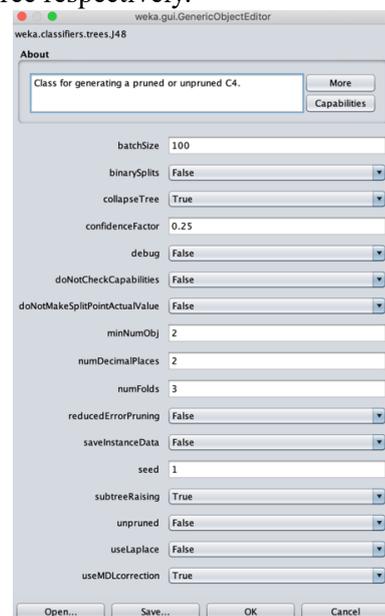

Figure 5: (a) StringToWordVector      (b) J48 Decision Tree

# 6 Evaluation

After the model was trained with the training data, labels for the test results were predicted from it. For predicting labels for the test data, output predictions option needs to be chosen which comes from the dialogue box when 'More options…' button is pressed, placed in the 'Test Options' box. The following screenshot of 'Test Options' box shows the 'More options…' button. CSV file was chosen from the list of options that appeared after selecting the output predictions. The text box where csv is written was clicked to open up the dialogue box for choosing the attributes for outputting the test labels. An empty csv file was created and it was selected in the output file options. '1' was given as input in the attributes option so that uid of the tweet can also be printed as output along with the test labels in the output csv file. All the other options were kept as default. The following image shows the screenshot of the options selected for outputting the test labels in the csv file.

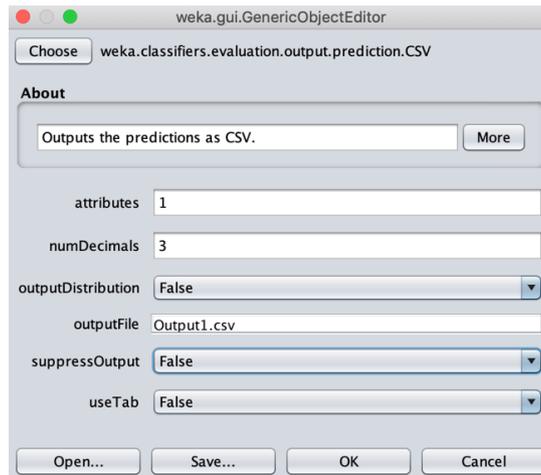

Figure 6: Output Predictions.

## 6.1 Evaluation Score

The csv file generated with the output predictions of the test data was made in the appropriate format required to be uploaded on the competition's website. The format required for uploading the results on the CodaLab website was Uid,Sentiment written in the first line followed by all the tweets' uid along with their sentiment polarity separated by comma written in separate lines. The output file needed to be renamed as answer.txt and to be uploaded as a zip file.

The output predictions on the test file was made twice as the model was trained two times. First training file contained 2323 tweets and the second training file contained 4591 tweets. After uploading the test labels file on the competition's website, F1-score was generated for the two files which was 0.4972 for the first file and 0.5316 for the second file. The following screenshot shows the F1-score obtained for the two files uploaded.

| # | SCORE | FILENAME | SUBMISSION DATE | STATUS | |
|---|---|---|---|---|---|
| 1 | --- | answer.txt.zip | 03/07/2020 23:15:47 | Failed | |
| 2 | --- | answer.zip | 03/08/2020 00:00:27 | Failed | |
| 3 | 0.497276 | answer.zip | 03/08/2020 00:05:24 | Finished | |
| 4 | 0.531685 | answer.zip | 03/08/2020 11:27:43 | Finished | ✓ |

Figure 7: Evaluation Score.

# References

CodaLab. 2019. *Sentimix Hindi-English.* [Online]. [Accessed 10 February 2020]. Available from: https://competitions.codalab.org/competitions/20654#learn_the_details-terms-and-condition.

MonkeyLearn. 2020. *Sentiment Analysis.* [Online]. [Accessed 18 June 2020]. Available from: https://monkeylearn.com/sentiment-analysis/#:~:text=Sentiment%20analysis%20is%20the%20interpretation,in%20online%20conversations%20and%20feedback.

Patwa, P., Aguilar, G., Kar, S., Pandey, S., Pykl, S., Garrette, D., Gambck, B., Chakraborty, T., Solorio, T. & Das, A. 2020. SemEval-2020 Sentimix Task 9: Overview of SENTIment Analysis of Code-MIXed Tweets. *Proceedings of the 14th International Workshop on Semantic Evaluation (SemEval-2020)*.

Sefik Ilkin Serengil. 2018. *A Step By Step C4.5 Decision Tree Example*. [Online]. [Accessed 15 July 2020]. Available from: https://sefiks.com/2018/05/13/a-step-by-step-c4-5-decision-tree-example/.

Sumit Saha. 2018. *What is the C4.5 algorithm and how does it work?* [Online]. [Accessed 15 July 2020]. Available from: https://towardsdatascience.com/what-is-the-c4-5-algorithm-and-how-does-it-work-2b971a9e7db0.

Wikipedia Contributors. 2020. *C4.5 algorithm*. [Online]. [Accessed 15 July 2020]. Available from: https://en.wikipedia.org/wiki/C4.5_algorithm.

Wikipedia contributors. 2020. *Weka (Machine Learning)*. Wikipedia, The Free Encyclopedia. [Online]. [Accessed 6 July 2020]. Available from: https://en.wikipedia.org/wiki/Weka_(machine_learning).